\documentclass{llncs}
\usepackage{epsfig}
\usepackage{remreset,aliascnt,llncsdoc}
\usepackage{amsmath,amssymb, mathrsfs}
\usepackage{times}
\usepackage{t1enc}
\usepackage{epstopdf}
\usepackage{multirow,array}
\usepackage{slashbox}
\usepackage{tablefootnote}

\title{Empirical Analysis of Sampling Based Estimators for Evaluating RBMs}
 \author{Vidyadhar Upadhya \and P. S. Sastry}
 \institute{Indian Institute of Science, Bangalore, India}
\psfull

\newcommand{\beq}{\begin{equation}}
\newcommand{\eeq}{\end{equation}}
\newcommand{\beqa}{\begin{eqnarray}}
\newcommand{\eeqa}{\end{eqnarray}}
\newcommand{\ben}{\begin{enumerate}}
\newcommand{\een}{\end{enumerate}}

\newcommand{\Real}{\mathbb{R}}

\newcommand{\lb}{\left(}
\newcommand{\rb}{\right)}
\newcommand{\ls}{\left[}
\newcommand{\rs}{\right]}

\newcommand{\vecx}{\mathbf{x}}

\newcommand{\vecv}{\mathbf{v}}

\newcommand{\vecb}{\mathbf{b}}
\newcommand{\vecc}{\mathbf{c}}
\newcommand{\vecw}{\mathbf{w}}
\newcommand{\vech}{\mathbf{h}}

\newcommand{\dow}{\partial}

\newcommand{\Ex}{\mathbb{E}}

\newcommand{\non}{\nonumber}

\begin{document}
\maketitle
 \begin{abstract}
The Restricted Boltzmann Machines (RBM) can be used either as classifiers or as generative models. 
The quality of the generative RBM is measured through the average log-likelihood on test data. 
Due to the high computational complexity of evaluating the partition function, exact calculation of 
test log-likelihood is very difficult. In recent years some estimation methods are 
suggested for approximate computation of test log-likelihood. In this paper we present an empirical 
comparison of the main estimation methods, namely, the AIS algorithm for estimating the partition 
function, the CSL method for directly estimating the log-likelihood, and the RAISE algorithm that 
combines these two ideas. We use the MNIST data set to learn the RBM and then compare these methods for estimating the 
test log-likelihood.
 \end{abstract}

\section{Introduction}\label{sec:intro}
The Restricted Boltzmann Machines (RBM) are among the basic building blocks of 
deep learning models. They can be used as classifiers as well as 
 generative models. The parameters of an RBM are learnt with an objective of maximizing the 
log-likelihood using, e.g., the contrastive divergence \cite{hinton2002training}.
The quality of the learnt generative RBM is 
evaluated based on the average log-likelihood which, for N test samples, is  
given by,
\beq
\mathscr{L}= \frac{1}{N}\sum_{i=1}^N \log p(\vecv^{(i)}) 
\eeq
where $p(\vecv^{(i)})$ is the probability (or the likelihood) of the 
$i^{\text{th}}$ test sample, $\vecv^{(i)}$. The model with higher average 
test log-likelihood is better. 
The test log-likelihood can also be used in devising a 
stopping criterion for the learning and for fixing the 
hyper-parameters through cross validation.

The likelihood $p(\vecv)$ can be written as $p^{*}(\vecv)/Z$. 
While $p^{*}(\vecv)$ is easy to evaluate, the normalizing
constant $Z$, called the partition function, is computationally expensive. 
Therefore, various sampling based estimators 
have been proposed
 for the estimation of the log-likelihood.
 In this paper we present an empirical analysis of the two main approaches 
for estimating the log-likelihood. 

The first approach of estimating the average test log-likelihood is to 
approximately estimate the partition function of the model distribution and then 
use it to calculate the log-likelihood. The partition function can be estimated 
with the Monte Carlo method using the samples obtained from the model 
distribution. However, generating independent samples is difficult since the model distribution 
which the RBM represents is complicated and high dimensional in most applications.
A useful sampling technique in such case is the importance sampling where
samples obtained from a simple distribution, called the proposal distribution, are used to estimate the partition function.
However, for large models, the variance of this estimator is very high and 
may even be infinite \cite{mackay2003information} when the proposal distribution is 
not a good approximation of the target distribution. To overcome this 
difficulty of choosing a good proposal distribution,  
a sequence of intermediate distributions is used. The annealed importance 
sampling (AIS) based estimator is one such estimator \cite{neal2001annealed}.

The second approach is to estimate the average test log-likelihood directly by  
marginalizing over the hidden variables from the model distribution, because the 
conditional distribution of the visible units given the hidden units is simple to evaluate in an RBM. 
However, the computational complexity grows exponentially with the number of 
hidden units present in the model. Also, this computation has to be 
repeated for each test sample. Hence, an approximate method which uses a 
sample based estimator called conservative sampling based likelihood estimator (CSL) is proposed in \cite{bengio2013bounding}. A 
more efficient method called reverse annealed importance sampling estimator 
(RAISE) implements CSL by formulating the problem of marginalization as a 
partition function estimation problem.

In this paper, we present an empirical comparison of the performance of the 
sampling based estimators, namely, annealed importance sampling estimator(AIS) 
\cite{neal2001annealed}, 
conservative sampling based likelihood estimator (CSL) 
\cite{bengio2013bounding}, 
reverse annealed importance sampling estimator (RAISE)\cite{burda2014accurate}.
Since our main aim is to evaluate the learnt RBM, the issues associated with the learning are 
not addressed in this paper.
Initially we learn RBM models (with different number of hidden units) using the standard CD algorithm on 
the MNIST dataset with suitably chosen hyperparameters. We evaluate the average test log-likelihood
of each of the learnt models using the estimators mentioned above and compare their performance. 
For models with 
small number of hidden units we can calculate the test log-likelihood through the brute force approach. When the ground truth is known we can readily evaluate the performance 
of different estimators. However, for large models we do not know the ground truth. 

The rest of the paper is organized as follows. In section \ref{sec:background} 
we first briefly describe the RBM model. We then explain the 
problem of computing the average test log likelihood for an RBM and explain the 
two approaches used for solving it. We then describe 
the different sampling methods considered in this study in section 
\ref{sec:sampling_method}. In section \ref{sec:Experiments} we describe the 
simulation setting, the parameters used in different estimators and then present 
results of our empirical study. Finally we conclude the paper in 
section~\ref{sec:conclusions}.

\section{Restricted Boltzmann Machines}
\label{sec:background}
The Restricted Boltzmann Machine (RBM) is a special case of the Boltzmann 
Machine where the intra-layer connections are restricted 
\cite{smolensky1986information,freund1994unsupervised,hinton2002training}. The 
visible stochastic units $(\vecv)$ are connected to the  hidden 
stochastic units $(\vech)$ through bidirectional links. We restrict our study to the binary RBM with   
$\vecv\in\{0,1\}^m$ and $\vech\in\{0,1\}^n$, being  
the states of visible and hidden units respectively. Note that,
$m$ and $n$ are the number of visible and hidden units. 
For the binary RBM, the energy function 
and  the probability that the model assigns to $\{\vecv,\vech\}$  is given by,
 \beqa
 E(\vecv,\vech;\theta)&=&-\sum_{i,j}w_{ij} h_i\, v_j-\sum_{j=1}^{m} 
b_j\,v_j-\sum_{i=1}^{n} c_i\, h_i\\
p(\vecv,\vech\vert\theta) &=&\frac{e^{-E(\vecv,\vech;\theta)}}{Z}
 \eeqa
 
 where $Z=\sum_{\vecv,\vech}e^{-E(\vecv,\vech;\theta)}$ is the partition 
function and $\theta=\{\vecw\in\Real^{n\times 
m},\vecb\in\Real^{m},\vecc\in\Real^{n}\}$ is the set of model parameters. 

\subsection{Evaluation of the Average Test Log-likelihood  for an RBM}
 As mentioned in section \ref{sec:intro} the  parameters of an RBM are learnt 
using the contastive divergence algorithm. Once the model is learnt, for a particular 
test data $\vecv$ the log-likelihood can be calculated as 
(see \cite{fischer2012introduction} for details),
\beqa
& \log &p(\vecv)=\log \sum_\vech e^{-E(\vecv,\vech;\theta)} -\log Z\non\\
&=&\sum\limits_j b_j v_j+\log \sum_\vech e^{\sum\limits_i c_i h_i+\sum\limits_j 
w_{ij} h_i v_j}-\log Z\non\\
&=& \sum\limits_j b_j v_j+\sum_i \log (1+e^{c_i+\sum\limits_j w_{ij} v_j})-\log 
Z\label{logpv_eq}
\eeqa

The difficulty in evaluating the above equation is due to the presence of 
intractable log partition function. The log partition function can be expanded 
as,
\beqa
\log Z &=& \log \sum_{\vecv,\vech}e^{-E(\vecv,\vech;\theta)}\non\\
&=&\log \sum_{\vecv} e^{\sum\limits_j b_j v_j} \sum_\vech e^{\sum\limits_i c_i 
h_i+\sum\limits_j w_{ij} h_i v_j}\non\\
&=&\log \sum_{\vecv} e^{\sum\limits_j b_j v_j} \prod_i \lb 
1+e^{c_i+\sum\limits_j w_{ij} v_j}\rb\label{part_fn_exp_v}
\eeqa
Similarly, it can be written in terms of $\vech$ as,
\beq
\log Z=\log \sum_{\vech} e^{\sum\limits_i c_i h_i} \prod_j \lb 
1+e^{b_j+\sum\limits_i w_{ij} h_i}\rb \label{part_fn_exp_h}
\eeq
The above equation can also be written in terms of $\vech$. Hence, to evaluate the partition function by using either eq.\eqref{part_fn_exp_v} or eq.\eqref{part_fn_exp_h} we 
have to sum $2^L$ terms, where $L=\min\{m,n\}$, which is computationally 
expensive for large models. However, if we can estimate the partition 
function by some other method then we can efficiently estimate the log-likelihood using eq.\eqref{logpv_eq}.


The test log-likelihood can also be estimated directly by 
marginalizing the model distribution over the latent variable $\vech$ through a sample average as,
\beq
\log p(\vecv)=\log \sum_\vech p(\vech) 
p(\vecv\vert\vech)\approx\log \frac{1}{N}\sum_{i=1}^N p(\vecv\vert\vech^{(i)})\label{eq:CSL_avg}
\eeq
 where $N$ samples of $\vech$ are obtained as $\vech^{(i)}\sim p(\vech),  \forall i$.

\section{Sampling Based Estimators for Evaluating RBMs}\label{sec:sampling_method}
In this section we briefly discuss various sampling based estimators for the test log-likelihood. 
As mentioned earlier, we either need a method to estimate $Z$, 
so that we can use eq.\eqref{logpv_eq}; or we need a method to sample 
from $p(\vech)$ so that we can use eq.\eqref{eq:CSL_avg} for evaluating the test log-likelihood. 
We discuss
\subsection{Annealed Importance Sampling for Estimating Partition Function}\label{sec:AIS}
Suppose two distributions $f_A(\vecx)=f^*_A(\vecx)/Z_A$ and 
$f_B(\vecx)=f^*_B(\vecx)/Z_B$ are given such that $f_A(\vecx)\ne 0$ when 
$f_B(\vecx)\ne 0$ and it is possible to obtain independent samples from 
$f_A(\vecx)$. It is also assumed that $f^*_A(\vecx)$, $f^*_B(\vecx)$ and $Z_A$ are easy to evaluate. Then the ratio of partition functions can be 
written as,
 \beq
  \frac{Z_B}{Z_A}=\int \frac{f^*_B(\vecx)} {Z_A} d\vecx = \int 
\frac{f^*_B(\vecx)}{f^*_A(\vecx)} f_A(\vecx)\, d\vecx\approx \frac{1}{M}\sum_{i=1}^M \frac{f^*_B(\vecx^{(i)})}{f^*_A(\vecx^{(i)})}\approx \frac{1}{M}\sum_{i=1}^M w_{\text{imp}}^{(i)}\label{eq:imp_samp}
  \eeq
 where $w_{\text{imp}}^{(i)}=f^*_B(\vecx^{(i)})/f^*_A(\vecx^{(i)})$ is termed as the $i^{\text{th}}$ importance weight and $\vecx^{(i)}$ are independent samples from the distribution $f_A$. However, the variance of this estimator is very high unless the proposal 
distribution, $f_A$, is a good approximation of the target distribution, $f_B$  
\cite{mackay2003information}. 
Hence, unless we can find a proposal distribution, $f_A$, that is easy to sample from, and at the same time, it is close to the target distribution, $f_B$, the importance sampling method does not 
give a good estimate.
In order to overcome this issue with choosing a good proposal 
distribution, a sequence of intermediate probability distributions is introduced 
to assist in moving gradually from the proposal to the target distribution 
\cite{neal2001annealed}. This method is called annealed importance sampling and 
it is the state of the art estimator for the partition function. 

Suppose there are $K-1$ intermediate distributions, $f_A=f_0,f_1,\ldots,f_{K-1},f_{K}=f_B$. As earlier, let each distribution be given by $f_i(\vecx)=f^*_i(\vecx)/Z_i, \,i=0,1,\ldots,K$. Then using eq.\eqref{eq:imp_samp}, we have, for every $k$, $0\le k\le K-1$,
\beq
\frac{Z_{k+1}}{Z_{k}}\approx \frac{1}{M}\sum_{i=1}^{M} 
\frac{f_{k+1}(\vecx_k^{(i)})}{f_{k}(\vecx_k^{(i)})},\text{ where } \vecx_k^{(i)}\sim 
f_k\label{ais_ratio_Mrun}
\eeq
If we can ensure that the $f_k$ and $f_{k+1}$ are close to each other then the above is a good estimate of the ratio of the partition functions. Further, we can write,
\beq
\frac{Z_K}{Z_0}=\frac{Z_1}{Z_0}\frac{Z_2}{
Z_1}\ldots\frac{Z_K}{Z_{K-1}}=\prod_{k=0}^{K-1}\frac{Z_{k+1}}{Z_{k}}\label{ais_ratio_inter_function}
\eeq
Since $f_0=f_A$ is a simple distribution, we can calculate $Z_0$ and hence the above gives us a good estimate for $Z_K$.

The next question is, how does one sample from each of the $f_k$, $0\le k\le K-1$?. A standard method for this is as follows. Initally a sample $\vecx_1$ is drawn from $f_0(\vecx)=f_A(\vecx)$. Then
for $1\le k \le K-1$,  the $(k+1)^{\text{th}}$ sample, $\vecx_{k+1}$, is obtained by sampling from  $T_k(\vecx_{k+1}\vert \vecx_{k})$
where $T_k$ is a transition function of a (reversible) Markov chain for which $f_k$ is the invariant distribution. This sequence of states, $\vecx_1,\vecx_1,\ldots\vecx_K$ is one sample to estimate $Z_k$ using eq.\eqref{ais_ratio_Mrun} and eq.\eqref{ais_ratio_inter_function}. Then we can get M samples by starting in an initial state sampled from $f_A$ 
and then going through the $K$ Markov transitions, $M$ times.

The key idea in this method is that by moving gradually from the proposal to the target distribution, the variance of importance weights can be kept small if
the two consecutive intermediate distributions differ by a small amount. For the detailed analysis and conditions for this estimator to work, refer to \cite{neal2001annealed}. 

The choice of intermediate probability distributions is problem specific. The 
geometric average of $f_A(\vecx)$ and $f_B(\vecx)$ is a popular choice in literature. Thus we keep,
\beq
f_k(\vecx)\propto f_A(\vecx)^{1-\beta_k} 
f_B(\vecx)^{\beta_k} \label{eqAIS}
\eeq
where $\beta_k$\textquoteright s are chosen such that  $0=\beta_0 
<\ldots<\beta_K=1$. 
 
 The following two facts give some insight about the partition function estimate 
obtained using the AIS estimator \cite{burda2014accurate}.  
The AIS estimator yields unbiased estimate of $Z$. However,  the Jensen's 
inequality shows that, on the average, AIS estimator underestimates the log partition function,
\beq
\Ex \ls\log \hat{Z}\rs\le \log \Ex\ls \hat{Z}\rs=\log Z\label{eq:AIS_bound}
\eeq
The underestimate of the log partition function results in overestimate for the 
test log-likelihood (see eq.\eqref{logpv_eq}). Then it is difficult to infer whether the model is good or we got an optimistic likelihood 
estimate.
However, the Markov inequality shows that overestimating the log partition 
function by a large amount is not very likely, 
\beq
P\ls\log \hat{Z}>\log Z+k\rs=P\ls\hat{Z}>e^{\log Z+k}\rs<\frac{\Ex\ls\hat{Z}\rs}{Z\,e^k}= e^{-k}
\eeq
 
\subsubsection{Estimating the Partition Function of an RBM:}
The partition function of an RBM can be evaluated using the AIS estimator 
\cite{salakhutdinov2008quantitative}. Consider the two RBMs with 
parameters $\theta^A=\{\vecw^A,\vecb^A,\vecc^A\}$ and $\theta^B=\{\vecw^B,\vecb^B,\vecc^B\}$ corresponding to the proposal distribution $(f_A)$ and the target distribution $(f_B)$ respectively.   
The proposal RBM distribution is chosen such that obtaining independent samples from it, is simple. The objective is to estimate the partition function of the RBM given by the parameter $\theta^B$, which is nothing but the set of model parameters of the learnt RBM.  

The two choices for the proposal RBM used in the 
present study are the \emph{uniform} and the \emph{base rate} RBM. The uniform 
distribution as a proposal can be implemented by taking 
$\vecw^A=0,\vecb^A=0,\vecc^A=0$. If we consider all zero weight matrix and
nonzero biases the proposal distribution is given as,
\beq
f_A(\vecv)=\frac{e^{\sum_j b_j v_j}}{\prod\limits_j \,1+e^{-b_j}}
\eeq
The ML solution for $b_j$ (solving  $\frac{\dow f_A}{\dow b_j}=0$) turns out to be 
$\log(\bar{\vecv})-\log(1-\bar{\vecv})$ where $\bar{\vecv}$ is the mean of the 
training samples. This is is termed as the \textquoteleft base rate\textquoteright  
RBM.

The intermediate distributions are constructed using the geometric average path. It corresponds to averaging the energy 
function $E(\vecv,\vech^A;\theta^A)$ of the proposal RBM and $E(\vecv,\vech^B;\theta^B)$ of the target RBM, i.e., 
\beq
E_k(\vecv,\vech){=}-\lb(1-\beta_k)\,E(\vecv,\vech^A;\theta^A)+ 
\beta_k\,E(\vecv,\vech^B,\theta^B)\rb\label{eq:AIS_energy}
\eeq
where  $E_k(\vecv,\vech)$ denotes the energy function for the intermediate distribution $f_k$ and $\vech$ denotes the stacked hidden units $\{\vech^A,\vech^B\}$.
The $k^\text{th}$ intermediate distribution is given by,
\beq
f_k(\vecv)= \frac{f_k^*(\vecv)}{Z_k}=\frac{1}{Z_k}\sum\limits_\vech 
e^{-E_k(\vecv,\vech)}\label{eq:fkv}
\eeq

Since each of the intermediate distributions $f_k$ is defined by the RBM, the samples are obtained from the invariant distribution through Gibbs sampler. More details on the Gibbs sampler implementation and the exact form of the intermediate distributions are available in \cite{salakhutdinov2008quantitative}.

The algorithm works as follows. Initially, a sample $\vecv_1$ is generated from the proposal RBM. Then, for $1\le k \le K-1$, 
the $(k+1)^{\text{th}}$ sample $\vecv^{(k+1)}$ is to be generated by $f_k$ (given in eq. eq.\eqref{eq:fkv}). This is done by sampling from 
$T_k(\vecv^{(k+1)}\vert \vecv^{(k)})$ where $T_k$ refers to the transition operator for which $f_k$ is the invariant distribution. Since $f_k$
is naturally defined for an RBM, the $T_k(\vecv^{(k+1)}\vert \vecv^{(k)})$ corresponds to Gibbs ampler for the RBM whose energy function is $E_k$ (given in eq. eq.\eqref{eq:AIS_energy}).
Once the samples are obtained, the ratio in eq.\eqref{ais_ratio_inter_function} is used to calculate the partition function.

\subsection{Conservative Sampling Based Log-likelihood Estimator}\label{sec:CSL}
Given the conditional probability $p(\vecv\vert\vech)$ and a set of samples $\{\vech^{(1)},\vech^{(2)}\ldots\vech^{(N)}\}$ 
from the distribution $p(\vech)$, 
the log-likelihood on a test sample using the CSL estimator is given in eq.\eqref{eq:CSL_avg}.

The hidden units samples are obtained using the Gibbs sampler for the RBM.
The state of the visible units can be initialized randomly (called the unbiased CSL) or by using the training samples (called the biased CSL). 
Then, the hidden units and the visible units are sampled alternately from the corresponding 
conditional distributions for a large number of iterations. 

The Jensen's inequality shows that the CSL estimator underestimates the test 
log-likelihood \cite{bengio2013bounding}, i.e.,
\beq
\Ex_\vech [\log \hat{p(\vecv})]\le \log \Ex_\vech [p(\vecv\vert \vech)]=\, \log p(\vecv)
\eeq

\subsection{Reverse AIS Estimator}\label{sec:RAISE}
The RAISE estimates the log-likelihood similar to the CSL approach while implementing it through the AIS approach \cite{burda2014accurate}. 
Consider the likelihood on a given test data $\vecv$ as in eq.\eqref{eq:CSL_avg}, 
$p(\vecv)=\sum_\vech p(\vech)p(\vec v\vert\vech)$.
Finding this $p(\vecv)$ is equivalent to finding the partition function of a distribution given by $f(\vech)=p(\vech)p(\vecv\vert\vech)$ for a fixed $\vecv$. Hence, the AIS estimator (with suitable modifications) can be used. For the analysis and the implementation details of the method refer \cite{burda2014accurate}.
 The brief description of the algorithm is given below.

The test data $\vecv$ is assumed to be a sample from the target distribution. This test sample is used as the initial state of 
the AIS Markov chain which is defined to move gradually from the target to the proposal distribution as opposed to moving from
the proposal to the target distribution as in the AIS estimator discussed in section \ref{sec:AIS}. 
The visible and hidden units, $\vech$ and $\vecv$, of intermediate distributions are sampled alternately from this chain using the 
transition operators similar to the ones defined for the AIS estimator.   
 Once the samples are obtained the estimate of the likelihood is given as \cite{burda2014accurate},
\beq
\hat{p(\vecv)} =\frac{f_{K}(\vecv)}{Z_0} \,\,\prod\limits_{k=1}^{K} \frac{f_{k-1}(\vecx_k)}{f_k(\vecx_k)}\label{eq:revais}
\eeq

\section{Experimental Results and Discussions}\label{sec:Experiments}
In this section we present the results of estimating the partition function and the average test log-likelihood using the different methods discussed in the previous section. 
We first learn the RBMs with different number of hidden units ($20$, $200$ and $500$, denoted as RBM$20$, RBM$200$, RBM$500$) using the standard CD-$20$ algorithm on the MNIST dataset.
We fix the learning rate, $\eta=0.1$, batch size $=100$, weight decay $=0.001$, initial momentum $=0.5$, final momentum $=0.9$ and change the momentum at $5^{\text{th}}$ step.
Initial weights are sampled from $U[-1/\sqrt{L},1/\sqrt{L}]$ where $L=\text{min}\{m,n\}$.

Then, for each of the learnt RBMs, we estimate the test log-likelihood on the MNIST test dataset using the different estimators. 
Note that, even though learning the RBM with different hyperparameter settings produce (significantly) different models,
we have observed that the behaviour of estimators are similar on these models. Therefore we present results for one such RBM which is learnt using the above hyperparameter settings.
For the case of RBM$20$, the ground truth is calculated by summing over all $2^{20}$ states so that we 
can assess accuracy of the estimates. 

\subsection{AIS}
The performance of the AIS estimator depends on the distribution of the proposal RBM, the number of intermediate distributions ($K$) and the number of samples ($M$, also called AIS runs). We fix $M=500$ and use uniform or baserate as the proposal distribution.   
The value of $K$ is chosen from $\{100,1000,10000\}$ and we use the linear path $\beta=[0:1/K:1]$ for all the experiments. The handcrafted schedule having $K=14500$ given in \cite{salakhutdinov2008quantitative}  with $\beta=[0:1/1000:0.5, 0.5:1/10000:0.9, 0.9:1/100000:1]$ is also implemented for comparison. 
In order to estimate the variance of the estimator we repeat the experiment $50$ times with a random initial state each time. 
Table \ref{table:h20_uniform_baserate_AIS} gives the estimate of $\log\,Z$, $\mathscr{L}$ and $\sigma$ (the standard deviation of the estimate of $\hat{\log \,Z}$).
\begin{table}[ht!]                                                                                          
\centering                             
\caption{AIS estimate of $\mathscr{L}$, \textbf{Ground truth: }$\mathbf{\,\,\log \,Z= 230.61}$, $\mathbf{\mathscr{L}= -141.24}$ for \textbf{RBM}$\mathbf{20}$}
\label{table:h20_uniform_baserate_AIS}  
\renewcommand{\arraystretch}{1.4}                                                                           
\renewcommand{\tabcolsep}{0.2cm} 
\begin{tabular}{ll}
\begin{tabular}{|c|c|c|c|c|} 
\multicolumn{5}{c}{ \emph{Uniform} Proposal} \\ 
\hline                                                                                                      
 $n$ & $K$ & $\widehat{\log\, Z}$ & $\hat{\mathscr{L}}$ &  $\sigma$ \\
  \hline
  \multirow{3}{*}{$20$} & $1000$ & $229.2660$ & $-139.8933$  & $0.2905$ \\
  & $10000$ & $229.5245$ & $-140.1518$ & $0.2423$  \\
  & $14500$ & $229.8978$ & $-140.5250$ & $0.6095$  \\
  \hline
 \multirow{3}{*}{$200$} & $1000$ & $174.1720$ & $-111.2346$ & $0.6724$  \\
  & $10000$ & $174.7664$ & $-111.8290$  & $0.1528$ \\
  & $14500$ & $174.7424$ & $-111.8050 $ & $0.2623$  \\
    \hline
   \multirow{3}{*}{$500$}  & $1000$ & $173.4807 $ & $-114.8451$ & $0.3071$  \\
  & $10000$ & $173.4802$ & $-114.8446$  & $0.1207$ \\
  & $14500$ & $173.4567$ & $-114.8211$ & $0.1902$  \\ 
  \hline 
\end{tabular}    
\begin{tabular}{|c|c|c|}      
\multicolumn{3}{c}{\emph{Baserate} Proposal} \\  
\hline                                                                                                      
 $\widehat{\log\, Z}$ & $\hat{\mathscr{L}}$ &  $\sigma$ \\
  \hline 
  $230.5353$ & $-141.1626$ & $0.2114$   \\ 
   $230.6192$ & $-141.2465$  & $0.0629$ \\
  $230.6082$ & $-141.2355$ & $0.0431$  \\
  \hline 
    $173.5514$ & $-110.6140$ & $1.2344$  \\
    $174.6819$ & $-111.7445$ & $0.2408$\\
    $174.7512$ & $-111.8138$ & $0.1300$  \\
   \hline
    $172.2157$ & $-113.5801$  & $1.1597$\\
     $173.4767$ & $-114.8411$  & $0.2244$  \\
    $173.4240$ & $-114.7883$ & $0.1717$  \\
  \hline   
  
\end{tabular}   
\end{tabular}
\end{table}

\noindent Based on the ground truth available for RBM$20$, 
we observe that the AIS estimator, on the average, overestimates the test log-likelihood. The use of base rate proposal distribution gives slightly better estimate with less variance compared to using the uniform proposal distribution for the RBM$20$. However, for the RBM$200$ and RBM$500$ the proposal distribution has no significant effect on the estimated value though it affects the variance of the estimate.
On the whole, the linear annealing schedule with $K=10,000$ seems to perform as well as the hand-crafted annealing schedule 
with $K=14,500$. 

\subsection{CSL}
The samples required for the CSL estimator are obtained through the Gibbs sampler which alternately samples the hidden and the visible units. We ignore the first $B$ samples to allow burn-in and then collect samples after every $T$ (called 'Thin' parameter) steps, discarding the samples in between, to avoid correlation. 
We observed that the estimates obtained with a single chain is poor even if we run the Gibbs sampler for a large number of steps. 
This possibly indicates the poor mixing rate of the Gibbs chain. Therefore we experiment with many parallel Gibbs chains with different initial states.

We first simulate $S_M$ parallel Gibbs chains with $S_T$ steps for both the biased and the unbiased CSL. Under various values of the parameters $M$ and $T$, the samples of $\vech$ (to estimate the log-likelihood) are selected from these simulated chains.
 The experiment is repeated $100$ times to find the variance of the estimate, where for each experiment we select a random burn-in value and select chains randomly. We consider $S_M=5000$ and $S_T=25000$ for RBM$20$ and $S_M=2500$ and $S_T=50000$ for the other two RBMs. We consider both small and large burn-in setup for the biased CSL estimator. 

We fix the value of $T$ to $100$ and vary the number of chains, $M$, keeping the number of samples, $N$, 
constant and these results are presented in Table~\ref{table:h500_CSL}. We observe from the table that better estimates with less variance can be obtained by increasing the number of chains. Note that we are keeping the total number of samples fixed even when we vary the number of chains. Thus, for example, for RBM$500$, we get lower variance by having a total of $100,000$ samples from $1000$ chains rather than  having $200,000$ samples but from only $500$ chains. Thus having more chains also reduces the total computational effort (because we 
can do with less number of samples).  


The final estimates are close to ground truth in case of RBM$20$, even though the accuracy here is a bit poorer than that of AIS. 
For RBM$200$ and RBM$500$, the estimates differ by a large amount from those obtained with the AIS, 
even when $N$ is very large. This deviation is
smaller for the biased CSL estimates obtained with small burn-in than that of the unbiased CSL estimates and the biased CSL estimates obtained with large burn-in (refer Table. \ref{table:h500_CSL}). This may be due to high level of correlation among the 
 samples. Further, we observe that, for the RBM$20$, the biased CSL estimate gives a lower bound.

We also experimented with varying $T$ keeping $M$ and $N$ constant (not presented here). We found that the improvement in the accuracy of estimates obtained with higher values of $T$ is not very significant though it uses higher number of Gibbs steps.
when $N$ is large and a very small value of burn-in is used.
  \begin{table}[ht!]                                                                                          
\centering                             
\caption{CSL estimate of $\mathscr{L}$ for $T=100$ and for various values of $M$, keeping $N$ constant.  \textbf{Ground truth: }$\mathbf{\,\,\log \,Z= 230.61}$, $\mathbf{\mathscr{L}= -141.24}$ for \textbf{RBM}$\mathbf{20}$}
\label{table:h500_CSL}  
\renewcommand{\arraystretch}{1.4}                                                                           
\renewcommand{\tabcolsep}{0.2cm} 
\begin{tabular}{lll}
\begin{tabular}{|c|c|c|c|} 
\multicolumn{4}{l}{   a. Unbiased CSL} \\
\multicolumn{4}{c}{$n=500$} \\ 
\hline
 $N$ & $M$ & $\hat{\mathscr{L}}$ & $\sigma^2$ \\
\hline    
    \multirow{3}{*}{$100\times10^3$} & $250$ &  $-296.18$ & $5.90$\\
  & $500$ & $-284.25$ & $5.87$ \\
  & $1000$ & $-272.62$ & $3.92$ \\
  \hline
  \multirow{3}{*}{$200\times10^3$} & $500$ &  $-270.14$ & $4.56$ \\
  & $1000$ & $-257.29$ & $3.19$ \\
  & $2000$ &  $-244.87$ & $2.49$\\ 
  \hline
\end{tabular}
\begin{tabular}{|c|c|} 
\multicolumn{2}{c}{}\\
\multicolumn{2}{c}{$n=200$} \\ 
\hline
   $\hat{\mathscr{L}}$ & $\sigma^2$  \\
  \hline
    $-183.67$ & $2.31$ \\
  $-175.54$ & $1.65$ \\
  $-167.70$ & $1.45$ \\
  \hline
  $-172.41$ & $1.65$\\
  $-165.04$ & $1.01$\\
  $-158.76$ & $0.68$\\ 
  \hline
\end{tabular}
\begin{tabular}{|c|c|c|} 
\multicolumn{3}{c}{}\\
\multicolumn{3}{c}{$n=20$} \\ 
\hline
$M$ & $\hat{\mathscr{L}}$ & $\sigma^2$ \\
\hline    
     $500$ & $-153.83$ & $0.83$ \\
   $1000$ & $-150.39$ & $0.58$ \\
      $2000$ &  $-147.94 $ & $0.31$ \\
  \hline
   $1000$ & $-148.61$ & $0.35$ \\
   $2000$ &  $-146.31$ & $0.20$ \\ 
   $4000$ &  $-144.44$ & $0.09$ \\ 
  \hline
\end{tabular}\\

\begin{tabular}{|c|c|c|c|} 
\multicolumn{4}{c}{b. Biased CSL with large Burn-in} \\ 
\hline    
    \multirow{3}{*}{$100\times10^3$} & $250$ &  $-219.34$ & $4.64$\\
  & $500$ & $-208.12$ & $3.86$ \\
  & $1000$ & $-197.22$ & $2.17$ \\
  \hline
  \multirow{3}{*}{$200\times10^3$} & $500$ &  $-198.84$ & $2.61$ \\
  & $1000$ & $-188.32$ & $1.65$ \\
  & $2000$ &  $-179.69$ & $1.24$\\ 
  \hline
\end{tabular}
\begin{tabular}{|c|c|} 
 \multicolumn{2}{c}{} \\ 
   \hline    
    $-176.94$ & $1.89$\\
  $-169.26$ & $1.47$\\
  $-162.05$ & $0.92$\\
  \hline
  $-167.04$ & $1.56$\\
  $-160.11$ & $0.93$\\
  $-154.05$ & $0.54$\\ 
  \hline
\end{tabular}
\begin{tabular}{|c|c|c|} 
 \multicolumn{3}{c}{} \\ 
\hline    
     $500$ & $-151.18$ & $0.53$\\
   $1000$ & $-148.41$ & $0.37$\\
      $2000$ &  $-146.22 $ & $0.25$\\
  \hline
   $1000$ & $-147.17$ & $0.26$\\
   $2000$ &  $-145.27$ & $0.17$\\ 
   $4000$ &  $-143.52$ & $0.09$\\ 
  \hline
\end{tabular}\\

\begin{tabular}{|c|c|c|c|} 
\multicolumn{4}{c}{c. Biased CSL with small Burn-in} \\ 
\hline    
    \multirow{3}{*}{$100\times10^3$} & $250$ &  $-187.83$ & $2.52$\\
  & $500$ & $-178.12$ & $1.99$ \\
  & $1000$ & $-169.56$ & $1.50$ \\
  \hline
  \multirow{3}{*}{$200\times10^3$} & $500$ &  $-177.09$ & $1.89$ \\
  & $1000$ & $-168.82$ & $1.42$ \\
  & $2000$ &  $ -161.87$ & $1.45$\\ 
  \hline
\end{tabular}
\begin{tabular}{|c|c|} 
 \multicolumn{2}{c}{} \\ 
   \hline    
    $-166.91$ & $1.04$\\
  $-160.23$ & $0.98$\\
  $-154.03$ & $0.54$\\
  \hline
  $-159.31$ & $0.99$\\
  $-153.27$ & $0.59$\\
  $-148.01$ & $0.39$\\ 
  \hline
\end{tabular}
\begin{tabular}{|c|c|c|} 
 \multicolumn{3}{c}{} \\ 
\hline    
     $500$ & $-147.82$ & $0.22$\\
   $1000$ & $-145.54$ & $0.23$\\
      $2000$ &  $-143.74 $ & $0.20$\\
  \hline
   $1000$ & $-145.31$ & $0.19$\\
   $2000$ &  $-143.57$ & $0.14$\\ 
   $4000$ &  $-142.06$ & $0.13$\\ 
  \hline
\end{tabular}
   \end{tabular}
\end{table}

\subsection{RAISE}
The RAISE estimator requires implementation of the AIS chain for each test sample. This makes the estimator computationally expensive 
because the MNIST dataset contains $10,000$ test samples. 
Therefore, for the experiment we randomly select a subset of size $500$ ($50$ from each class) from the test dataset and then estimate the average test log-likelihood on this subset. We experiment with both uniform and base rate proposal distribution by fixing the number of AIS runs and varying the number of intermediate distributions, $K$. We also estimate the test log-likelihood on the chosen test subset using the AIS and CSL estimators, for comparison with the RAISE estimates. We keep the number of Gibbs steps used in the AIS and CSL estimators equal. 

We observe that, only when $K$ is very large and proposal distribution is uniform, the estimator provides conservative estimates. 
However, the baserate proposal gives overestimates even when $K=10000$. 
The lower bound on the test log-likelihood is similar to that of CSL estimate for RBM$20$.
However, unlike the case for CSL, for larger RBMs, the RAISE estimates matches closely with the AIS estimates. 

 \begin{table}[ht!]                                                                                          
\centering                             
\caption{RAISE estimate of $\mathscr{L}$ with $500$ test set, \textbf{Ground truth: }$\mathbf{\mathscr{L}= -142.39}$ for \textbf{RBM}$\mathbf{20}$. The baserate AIS with $K=10000$, $M=500$ is used. 
For the CSL, $5000\times 10^3$ samples are used to make the number of Gibbs steps equal to that of AIS.}
\label{table:h20_uniform_RAISE}  
\renewcommand{\arraystretch}{1.4}                                                                           
\renewcommand{\tabcolsep}{0.2cm}     
\begin{tabular}{ll}
\begin{tabular}{|c|c|c|c|}                     
\hline                                                                                                      
 $n$ & \backslashbox{Proposal}{$K$}  & $1000$ & $10000$  \\                                             
\hline 
\multirow{2}{*}{$20$} & Uniform & $-147.25$ & $-145.99$  \\
     & Baserate & $-146.97$ & $-144.14$  \\
\hline 
\multirow{2}{*}{$200$} & Uniform &  $-109.29$ & $-112.46$  \\
     & Baserate & $-110.42$ & $-109.01$  \\
\hline 
\multirow{2}{*}{$500$} & Uniform & $-114.75$ & $-118.02$  \\
     & Baserate  & $-108.75$ & $-112.04$  \\
\hline 
\end{tabular}    
&
 \renewcommand{\arraystretch}{1.45} 
\begin{tabular}{|c|c|}
 \hline
AIS & CSL\\
\hline
$-142.38$ & $-143.58$\\
& \\
\hline
$-112.96$ & $-142.64$\\
&\\
\hline
$-116.46$ & $-154.76$\\
&\\
\hline
\end{tabular}

\end{tabular}
\end{table}

\section{Conclusion}\label{sec:conclusions}
Calculating the average test log-likelihood of a learnt RBM is important for evaluating different learning strategies.   
In this paper we present extensive empirical analysis of the sampling based estimators for average test log-likelihood to gain 
insight into the performance of these methods. We experiment with RBMs with $20$, $200$ and $500$ hidden units. 
We observed that the  AIS estimator delivers good estimates with low variance. 
We also observe that the proposal distribution
does not seem to have much influence on the estimate.

Compared to the AIS estimate the CSL estimate is poorer and its variance is 
high especially for the RBM$200$ and RBM$500$. 
The estimated value also differ significantly with that of AIS. However, CSL is a
much simpler estimator computationally. 
We also showed that better estimates can be obtained with less computational effort by using multiple 
independent chains to generate samples. 
The biased CSL with a small burn-in provides the best estimate.

Unlike AIS, the RAISE gives conservative estimates. 
More importantly, for large RBMs, the deviation of RAISE estimate from the AIS estimate is not very large
compared to that of the deviation of CSL estimate from the AIS estimate. It means, for large $K$, the RAISE estimate will have tighter lower bound than the CSL estimate.
However it is computationally much more expensive. The conservativeness of RAISE may not be enough to justify the high computational cost.

Since large hidden unit RBMs are an important part of deep networks such as stacked-RBMs and DBNs, 
one needs efficient estimators to evaluate the learnt networks. Our empirical study indicates 
that there may be much scope for improving CSL like estimators to come up with 
computationally simple methods to get good estimates of the average test log-likelihood. 

\bibliographystyle{splncs03}
\bibliography{reference_proper}
\end{document}